\pdfoutput=1

\documentclass[11pt]{article}

\usepackage[]{EMNLP2023}

\usepackage{times}
\usepackage{latexsym}

\usepackage[T1]{fontenc}

\usepackage[utf8]{inputenc}

\usepackage{microtype}

\usepackage{inconsolata}
\usepackage{graphicx}
\usepackage{booktabs}

\title{Improving Dialogue Management: Quality Datasets vs Models}

\author{%
  Miguel Ángel Medina-Ramírez \\
  Siani\\
  ULPGC\\
  \footnotesize\texttt{miguelangel.medina@ulpgc.es} \\
  \And
  Cayetano Guerra-Artal \\
  Siani\\
  ULPGC\\
  \footnotesize\texttt{cayetano.guerra@ulpgc.es} \\
  \And
  Mario Hernández-Tejera \\
  Siani\\
  ULPGC\\
  \footnotesize\texttt{mario.hernandez@ulpgc.es}
}

\begin{document}
\maketitle
\begin{abstract}
Task-oriented dialogue systems (TODS) have become crucial for users to interact with machines and computers using natural language. One of its key components is the dialogue manager, which guides the conversation towards a good goal for the user by providing the best possible response. Previous works have proposed rule-based systems (RBS), reinforcement learning (RL), and supervised learning (SL) as solutions for the correct dialogue management; in other words, select the best response given input by the user. However, this work argues that the leading cause of DMs not achieving maximum performance resides in the quality of the datasets rather than the models employed thus far; this means that dataset errors, like mislabeling, originate a large percentage of failures in dialogue management. We studied the main errors in the most widely used datasets, Multiwoz 2.1 and SGD, to demonstrate this hypothesis. To do this, we have designed a synthetic dialogue generator to fully control the amount and type of errors introduced in the dataset. Using this generator, we demonstrated that errors in the datasets contribute proportionally to the performance of the models.
\end{abstract}

\section{Introduction}

TODS are a specialized Natural Language Processing (NLP) class designed to enable users to interact with computer systems to accomplish specific tasks. TODS represent a highly active research area due to their potential to improve human-computer interaction and provide users with seamless and efficient task completion. Recent advancements in Artificial Intelligence (AI) and Machine Learning (ML) have fuelled the proliferation of TODS and the exploration of novel architectures and techniques. One of the most widely used approaches due to its simplicity and controllability is the modular pipeline approach \cite{surveyMarcos2022, surveyNi2022, surveyZhang2020} as shown in the figure \ref{fig:chatbot_pipeline}. It consists of four modules:

\begin{figure}[!htbp]
\centering
\includegraphics[width=0.55\textwidth]{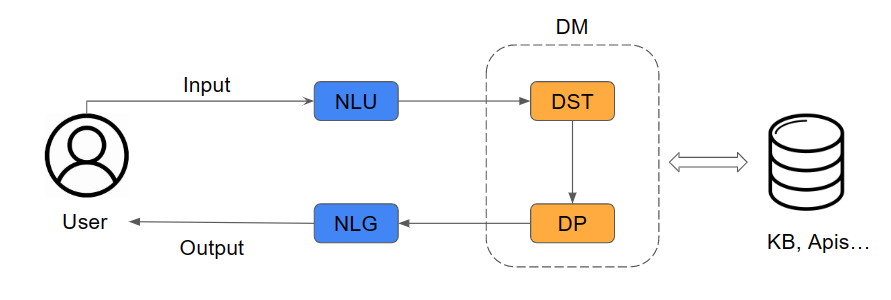}
\caption{Structure of a task-oriented dialogue system in the task-completion pipeline.}
\label{fig:chatbot_pipeline}
\end{figure}

\begin{itemize}
 \item \textbf{Natural Language Understanding (NLU)}: This module transforms the raw user message into  user intentions,  slots and domains. However, some recent modular systems \cite{deleteNLU2018} omit this module and use the raw user message as the input of the next module.

    \item \textbf{Dialogue State Tracking (DST)}: This module iteratively calibrates the dialogue states based on the current input and dialogue history. The dialogue state includes related user intentions and slot-value pairs.

    \item \textbf{Dialogue Policy Learning (DPL)}: Based on the calibrated dialogue states from the DST module, this module decides the following action of a dialogue agent.

    \item \textbf{Natural Language Generation (NLG)}: This module converts the selected dialogue actions into surface-level natural language, usually the ultimate response form.
    
\end{itemize}

DST and DPL are the components of Dialogue Managers (DM) in TODS. Rule-based solutions were initially utilized but faced limitations such as domain complexity and task scalability \cite{Weizenbaum1966}. With advancements in deep learning and the availability of labelled conversational datasets, supervised learning (SL) and reinforcement learning (RL) emerged as viable alternatives for training dialogue policies \cite{surveyNi2022, zhang-etal-2019-budgeted}. RL techniques have shown promise through optimizing dialogue policies via user interactions but still face challenges, such as the need for rule-based user simulators and domain-specific reward functions \cite{surveyZhang2020, surveyNi2022}. SL approaches, which involve the assignment of classified states to predefined system actions, have proven to be an excellent alternative to RL algorithms, as demonstrated in \cite{li-etal-2020-rethinking}; Researchers have proposed numerous models based on Transformers, GRU, LSTM, and multilayer perceptrons \cite{Vlasov2019, Vlasov2018REPD, li-etal-2020-rethinking, pepd2022}. However, the limited representativeness of available datasets may hinder supervised learning approaches, affecting the generalizability of learned policies and potentially requiring expensive data acquisition efforts.

As SL models must classify within a limited range of actions, the problem seems relatively straightforward. Nonetheless, none of the studied models in this work attains perfect accuracy. In this work, one of the main results is that the leading cause of the drop in performance is not mainly due to the models but to the data quality. Therefore, the data sets that evaluate these systems need to be correctly curated for a fair comparison. The objectives of this work are as follows:

\begin{enumerate}
    \item To analyse and elucidate the taxonomy of different errors found in datasets. To this end, we have studied Multiwoz 2.1 because \cite{Mosing2020context} performs an exhaustive analysis of this dataset, concluding that it presents several errors that affect performance.
    \item With this definition, we developed a synthetic dialogue generator that facilitates the creation of error-free and error-controlled datasets for research purposes. With this generator, we can control the number of dialogues, user intentions, entities and actions, and the probability of events in the dialogues, such as context switching or chit-chat, and we can especially control the probability and types of errors.
\end{enumerate}

We show that using curated datasets with this generator leads to superior performance, regardless of the architectural complexity of SL models, so our initial hypothesis that the problem is straightforward to handle has empirical evidence. Furthermore, introducing errors leads to a significant drop in performance, with a linear trend in all models. Therefore, this generator also serves as a tool to test the robustness of the models we are evaluating, making it a helpful tool.

\section{Related Work} 

Limited research focuses on studying and analysing datasets in the field of dialogue management in chatbots. However, recent works such as \cite{Mosing2020context} have examined the quality of datasets used in this field. This study's authors argue that many currently available datasets need more context and adequately reflect the complexity and diversity of human conversations. The authors evaluate the quality of these datasets using two popular datasets, multiwoz2.1 \cite{eric-etal-2020-multiwoz}, and Taskmaster-1 \cite{byrne-etal-2019-taskmaster}. Through a detailed analysis of these datasets, the authors identify various areas in which these datasets lack context, including history independence, solid knowledge base dependence, and ambiguous system responses. 

Other datasets, such as SGD \cite{rastogi2020sgd}  and multiwoz2.4 \cite{ye-etal-2022-multiwoz2.4}, have focused on improving existing datasets to solve different tasks. SGD \cite{rastogi2020sgd} presents a cleaner and more research-oriented dataset for agent generalisation. In contrast, multiwoz2.4 \cite{ye-etal-2022-multiwoz2.4} modifies the multiwoz2.1 dataset regarding slots and entities to improve dialogue state tracking performance. Other studies, such as \cite{deleteNLU2018}, suggest that the dialogue manager depends on NLU. 

Regarding dialogue generators, studies like \cite{grosz-sidner-1986-attention} suggest creating a dialogue generation by following a stack of topics. RavenClaw dialogue system \cite{Bohus2009ravenClaw} implemented this dialogue stack for handling sub-dialogues. However, while a stack structure effectively allows for the handling and conclusion of sub-dialogues, it can also be limiting. Ravenclaw's authors advocate for precise topic tracking to facilitate contextual interpretation of user intents. As human conversations often revisit and interleave topics, there is a need for a more flexible structure for an agent to handle dialogue.

Furthermore, one of the more flexible data structures is a graph, \cite{GrittaCG2021} proposes a method for improving the management of non-deterministic dialogues using a conversation graph that represents the possible responses and transitions between dialogue states. Besides, \cite{andreas-etal-2020-task} proposes a novel framework for task-oriented dialogue based on data flow synthesis, which involves transforming users' linguistic inputs into executable programs that manipulate data and external services. The authors represent the dialogue state as a data flow graph. Each node is a variable or an external service, and each edge is an operation or a connection. The dialogue manager maps each user input to a program that extends this graph with new nodes and edges. 

As we see in \cite{GrittaCG2021, andreas-etal-2020-task},  the graph is the most powerful data structure for dialogue generation. A good representation of a dialogue is a path in the conversational graph, where the nodes represent the current intentions and slots of the dialogue, and the edges represent the possible actions that the model can take based on the current and previous states.

\section{Synthetic dialogue generator}

Our primary motivation for creating the dialogue generator was that we needed curated datasets to which we could induce controlled errors to see how the presence or absence of errors affected the performance of the models. Therefore, we needed an algorithm or procedure that would allow us, on the one hand, to generate these synthetic sets and, on the other hand, to parameterise different aspects of this data. We have ruled out using generative models precisely because we want to generate symbolic code that represents intentions, actions and slots. Instead, we have designed an RBS because it offers superior controllability for our task than generative models. Furthermore, these procedures allow us to introduce randomisation mechanisms that can intentionally change the context or add errors. All these features are mod using configuration files, and this set of texts is called ontology.

\subsection{Ontology}

We define ontology as the information related to the set of actions, intentions, and slots required to satisfactorily achieve the various objectives of the dialogue. This ontology includes:

\begin{itemize}
    \item \textbf{Topic}: the set of slots belonging to a single domain. The bot must fill a set of slots by asking the user or providing possible values. There are three categories for slots:
    \begin{itemize}
        \item \textit{Mandatory}: Essential slots to complete the topic. They are either actively provided by the user or requested by the dialogue management module.
        \item \textit{Desired}:  Slots that are actively provided by the user or requested by the dialogue management module, but the task can still be completed if not filled in.
        \item \textit{Optional}: Unnecessary slots to complete the task. They are collected when the user provides but never explicitly requested by the dialogue management module.
    \end{itemize}
    \item \textbf{Domain}: A set of topics the chatbot is programmed to solve and their relationships. For example, in the domain of restaurants, two topics could be: finding a restaurant and ordering tickets for a concert.
\end{itemize}

Domains, topics, and slots are fully customisable. In the case of intention and actions, we create a simple map for many cases without any ambiguities.

\subsection{Intentions and Actions}

One of the essential aspects of the generator is the intentions and actions that represent what the user can say and the possible responses of the bot. Therefore, our approach was to make them as general as possible to cover many domains.

\begin{itemize}
    \item \textbf{Intentions}: Intentions are predefined actions that represent the motivations behind user queries. They are categories that encompass different types of requests and help the bot generate appropriate responses.:
    \begin{itemize}
        \item INFORM INTENT: The user indicates the task he/she wants to perform (e.g., to book a restaurant). There can be more than one in the input sentence (Example: I want to make a reservation at a restaurant, and I also want to order a taxi to take us there).
        \item INFORM: The user can inform the bot about the value of a single slot through the intention. The system will generate multiple.
        \item AFFIRM: Positive response to a bot query.
        \item NEGATE: Negative response to a bot query.
        \item REQUEST: The user asks for the value of a slot (Example: What kind of food did I ask you?).
        \item THANK: to show gratitude.
        \item GOODBYE: for goodbyes.
        \item UNK: for those entries that the NLU cannot classify.
        \item CHIT CHAT: for all those entries that deviate from the domains in the dataset.
    \end{itemize}

    \item \textbf{Actions}: Each of the possible responses of the bot to the current state of the dialogue. We can solve many scenarios with the following actions. However, there is a limit to the number of actions.

    \begin{itemize}
        \item INFORM: to inform or offer a slot to the registered or unregistered user.
        \item REQUEST: to request a mandatory slot from the user.
        \item CONFIRM: Confirm that the model registered the slot.
        \item NOTIFY: to notify the search status if it has succeeded.
        \item REQ MORE: to request a mandatory slot from the user.
        \item ANSWER CHIT CHAT: reply to the chitchat
    \end{itemize}
    
\end{itemize}

\subsection{Rules}

According to \cite{GrittaCG2021, andreas-etal-2020-task}, we seek to generate a graph for each data set, where the nodes are the states of the dialogue, composed of intentions, actions and slots, and the links are the corresponding actions. Each node will have information related to the domain and the corresponding topic. However, implementing this theoretical interpretation of a conversation graph can be challenging in practice due to the many different contexts and events that can change the path of the graph; the user can change their mind during a conversation, which can alter the course of the conversation. For instance, when ordering a pizza, the user may change their order based on their dietary preferences or decide to dine instead of placing a take-out order. We use the "stack of topics" proposed by \cite{Bohus2009ravenClaw}  as the next level of abstraction in a dialogue. We could jump into the context, change slots, or even chit-chat in a conversation. These events are hard to implement using a raw graph; however, we design these events as topics in a stack, so on the top, we process one path without knowing the complete graph a priory. The graph emerges from following the structure of the stack. As a generator, there are randomization mechanisms that can change the context or intentionally add errors. Our generator applies the rules at the top of the pile, adapting them to the node domain and topic. The obligatory slots are the aim of all dialogue-oriented tasks, and we design all rules according to this principle:

\begin{itemize}
    \item We have a corresponding slot for every INFORM intent, which allows the user to input information for an empty slot or modify the value of a filled slot.
    \item The INFORM INTENT intent will be the one that starts a dialogue and has no associated slot.
    \item The corresponding action for INFORM is CONFIRM.
    \item If any mandatory slots are missing, then the action will be REQUEST. 
    \item The NOTIFY action will occur once the dialogue fills all the slots, indicating that an external source has been searched or requested.
    \item The model will trigger the REQ MORE action once the user fills all the required slots.
    \item ANSWER CHIT CHAT will occur whenever the intent is CHIT-CHAT.
    \item At any time an event can occur that changes the top of the context stack, all information is stored to continue when the dialogue returns to the top of the stack.

\end{itemize}

\subsection{Events}

An event is any conversation that disturbs achieving the current objective at the top of the dialogue stack. So we could highlight three types of events:

\begin{itemize}
    \item \textbf{Chit chat}: any conversation that departs from the defined domains of the dataset. Always come with an intention-action pair: CHIT CHAT and ANSWER CHIT CHAT.

  \item \textbf{Mind-changing}: when we have a slot filled with a specific value, but the user changes his mind by changing its value or leaving it empty.

  \item \textbf{Domain-changing}: when the user wants to complete a task in a specific domain but changes the topic or domain at any given time.
\end{itemize}

\subsection{Errors}

Unfortunately, errors are inherent in creating any dataset and may be due to incorrect labelling or poor transcription. When designing a dataset, we need to consider the importance of cleaning our data and checking that all samples are appropriate for the problem we want to solve. In addition, the performance of the models will be directly affected by perturbations in the dataset. This lack of performance is due to the nature of supervised learning models. If we train the algorithms on low-quality samples, we cannot guarantee they will obtain a good generalization and correct score.

In this section, we study and analyze each of these errors in the data sets applied to TOD, which according to \cite{Mosing2020context}, are very present in many of these sets, mainly in Multiwoz2.1:

\begin{itemize}
    \item \textbf{NLU errors}: If the NLU model does not perform a good classification of the input text, the performance of the dialogue manager will be seriously affected, causing the conversation management to fail.
    \item \textbf{Human labelling errors}: The labeller (a person) has incorrectly labelled these samples. These errors can be a misallocation of tags to intentions, actions or slots.
    \item \textbf{Limited temporal reference}: Some algorithms, such as TED, are designed to capture temporal dependencies in long conversations. The idea behind this is that the manager needs long-term context information for a dialogue manager to take the right action in a conversation. While this idea may make sense, in reality, datasets are designed intentionally or out of ignorance, with only the previous state in mind; this would not be a problem if the solution one wishes to propose is Markovian; however, this is not the case in a conversation, and humans do not make decisions based solely on the previous state in a conversation. Thus, the poor temporal generalization of the datasets affects the models used in production, which need to be well-trained to handle such issues. This error is studied in depth by \cite{Mosing2020context}.
    \item \textbf{Ambiguities}: We have included this phenomenon as an error because it can cause a substantial performance drop in the models if not considered. In reality, it is an inherent ambiguity in human language. When analyzing a dataset, it is possible to find multiple actions for a given dialogue state that do not impact the overall outcome of the conversation. Conversations can take various valid and coherent paths to communicate the intended message effectively. Therefore, trained models using this data can take different actions for the same state that are correct. This one-to-many nature can confound many algorithms designed to obtain the best possible answer. A proposed solution by \cite{convlad2019} involves creating atomic actions to expand the action space. This method combines actions with one or more different slots to simplify the problem and improve model performance. We have utilized this method to train dialogue management models for both synthetic and real data.
\end{itemize}

In this work, we focus only on NLU and mislabelling errors, as they are the most common and abundant in a dataset and can control by probability. Perturbation techniques for the generator consist of choosing a random sample from the dataset, consisting of intentions, slots and actions, and replacing its actual value with one chosen randomly from all possible ones. Another technique is to replace its actual value with a "UNK" (unknown), pretending that the labeller failed to identify the sample or the NLU model did not classify it well. We can control these error mechanisms by parameters that independently simulate the probability of this happening for actions, intentions and slots.

\section{Experimental Setup}

To conduct our analysis, we selected two widely used datasets in academic literature: MultiWOZ2.1 \footnote{\url{https://huggingface.co/datasets/ConvLab/multiwoz21}} \cite{eric-etal-2020-multiwoz} and SGD \footnote{\url{https://huggingface.co/datasets/ConvLab/sgd}} \cite{rastogi2020sgd}, along with three synthetic datasets produced by our generator.  In the case of Multiwoz2.1, we had to neglect 1377 dialogues since their annotations are incomplete. The primary objective of our experiments is to compare the performance of state-of-the-art DPL models using real, partially curated datasets and synthetic, error-free datasets. This allows us to obtain empirical evidence regarding the impact of errors on model performance.

The experiments' secondary objective is progressively increasing the number of errors in the synthetic dataset and observing the corresponding drop in models performance. Detailed information regarding the datasets can be found in Table \ref{tab:dataset_summary}.

Regarding the probability of event occurrences, such as chit-chat, mind-changing, and domain-changing, we have arbitrarily set a value of 20\%. We tested higher and lower values, but they did not show a significant influence. 

This work evaluated action classification using three key metrics: the F1 score, precision, and recall. Chosen for their proven effectiveness in gauging the performance of classification models, these metrics offer a comprehensive overview of the model's performance. They encapsulate the model's capacity for accurate prediction, robustness in varied situations, and overall reliability in identifying correct actions within dialogues.

We utilized an NVIDIA GeForce RTX 3090 for all our experiments to carry out the computations, with all models being executed in less than 24 hours for all datasets.

\begin{table}
\centering
\scriptsize
\begin{tabular}{@{}lcccccc@{}}
\toprule
\textbf{}           & \multicolumn{2}{c}{\textbf{Real}}  & \multicolumn{3}{c}{\textbf{Synthetic}}              \\ \cmidrule(lr){2-3} \cmidrule(lr){4-6}
\textbf{}           & \textbf{MultiWoz2.1} & \textbf{SGD} & \textbf{Simple} & \textbf{Medium} & \textbf{Hard} \\ 
\midrule
\textbf{Dialogues}  & 10438                & 20000        & 2000            & 6000            & 10438           \\
\textbf{Domains}    & 7                    & 20           & 2               & 5               & 7               \\
\textbf{Actions}    & 26                  & 30          & 8              & 13              & 26              \\
\textbf{Train}      & 8438                 & 16000        & 1200            & 3600            & 8438            \\
\textbf{Val}        & 1000                 & 2000         & 400             & 1200            & 1000            \\
\textbf{Test}       & 1000                 & 2000         & 400             & 1200            & 1000            \\ 
\bottomrule
\end{tabular}
\caption{\textbf{Summary of datasets}: The datasets vary in terms of the number of dialogues, domains, and slots, providing different levels of complexity for training and testing conversational models. The table also indicates the number of dialogues allocated for training, validation, and testing.}
\label{tab:dataset_summary}
\end{table}

\subsection{Models}

We have utilized the most relevant models regarding dialogue management while keeping the same hyperparameter configurations as specified by their respective authors.

\begin{itemize}
    \item \textbf{Transformer embedding dialogues (TED)} \cite{Vlasov2019}  uses the Star-Space algorithm, developed by Facebook \cite{Wu_Fisch_Chopra_Adams_Bordes_Weston_2018}. TED's primary goal is to enhance chatbots' performance in dialogue tasks by employing transformer-based encoders to capture temporal relations in the dialogues.

    \item \textbf{Recurrent embedding dialogues (RED)} \cite{Vlasov2019}  is the same network as TED but uses an LSTM encoder \cite{jurgen1997lstm} rather than transformer-based encoders.

    \item \textbf{Planning Enhanced Dialog Policy (PEDP)}\cite{pepd2022}  improves the performance of chatbots in dialogue tasks by using a planning module to predict intermediate states and individual actions.

    \item \textbf{DiaMultiClass (MC)} \cite{li-etal-2020-rethinking} is a three-layer MLP.

    \item \textbf{DiaSeq (SEQ)} \cite{li-etal-2020-rethinking} is a two-layer perceptron to extract features from raw state representations and uses a GRU to predict the following action.

    \item \textbf{DiaMultiDense (MD)}\cite{li-etal-2020-rethinking}  uses a two-layer MLP to extract state features, followed by an ensemble of dense layer and Gumbel-Softmax \cite{jang2016gumbelsoftmax} functions consecutively.
\end{itemize}

\subsection{Dialogue State}

We formulate a state representation according to \cite{li-etal-2020-rethinking}. There are four main types of information in the final representation: (1) the current slots, (2) the last user intent, (3) the last system action, and (4) the current dialogue management state. We employ a standard state for RED and TED as proposed in \cite{Vlasov2019}. To create this representation, we employ a binary embedding that incorporates the aforementioned types of information.

This work considers the bot response problem a multi-label prediction task, allowing for combined atomic actions within the same dialogue turn. Each action consists of a concatenation of the domain name, action type, and slot name.

\subsection{Results and Discussion}

We evaluated different models using real datasets, Multiwoz 2.1 and SGD, and we present the results in Table \ref{tab:all_datasets}. In the Multiwoz 2.1 dataset, the RED model achieved the highest results in F1 and Recall, with both values at 69.52\%. On the other hand, the PEDP model stood out for its precision, which reached a maximum value of 78.11\%, suggesting that this model was particularly effective in minimizing false positive responses.

Alternatively,  in the SGD dataset, the SEQ model stood out, achieving the highest F1 and Recall values, at 86.04\% and 84.65\%, respectively. This reflects that the SEQ model provided the best performance in terms of balance between precision and recall in this dataset. However, it was the PEDP model that achieved the highest precision, with a value of 92.07\%, indicating that this model was extremely effective at generating correct positive predictions.  These results vary between the two datasets, underscoring that models can perform differently depending on the particular characteristics of the dataset they are working with. Overall, it appears that all models performed better with the SGD dataset compared to Multiwoz 2.1. In addition to evaluating the models with the real datasets Multiwoz 2.1 and SGD, we also conducted tests with synthetic data. These synthetic datasets were generated with different levels of complexity: simple, medium, and hard.

In the simple synthetic dataset, both the RED and SEQ models achieved perfection in all evaluation metrics, reaching 100\% in F1, Precision, and Recall. This indicates that both models were capable of handling this dataset with high precision and completeness. On the other hand, the TED, MD, MC, and PEDP models performed less well, although all achieved a good performance. As the complexity increased with the medium synthetic dataset, the SEQ model maintained its perfect performance. The RED model experienced a slight drop in performance, although it remained high. In contrast, the other models showed a similar performance to what was observed in the simple synthetic dataset. Finally, on the hard synthetic dataset, the SEQ model consistently demonstrated exceptional performance, achieving nearly 100\% in all metrics. The rest of the models showed a slight decrease in their performance compared to the less complex synthetic datasets, indicating that the increasing difficulty of the data poses additional challenges for these models.

\begin{figure}[!htbp]
\centering
\includegraphics[width=0.50\textwidth]{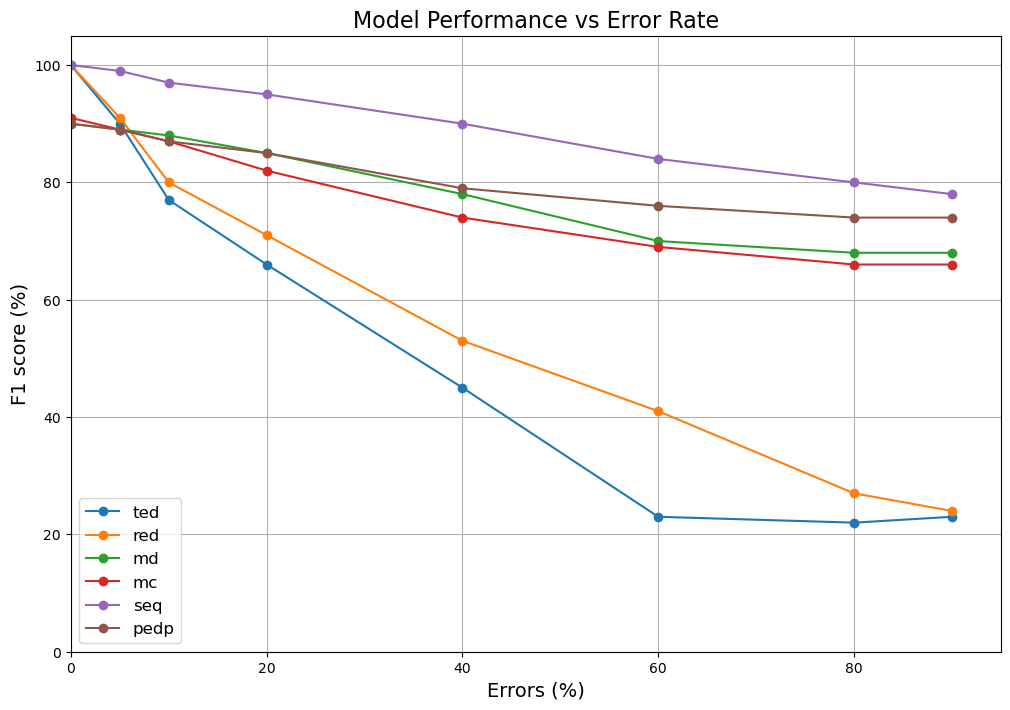}
\caption{The ability of systems to maintain their performance in the presence of NLU or labelling errors.}
\label{fig:robustnnes_NLU_erros}
\end{figure}

Continuing with the robustness tests of the models, we also explored how they behave in the presence of errors in the datasets. To do this, we gradually increased the proportion of errors in the synthetic datasets and observed its impact on the performance of the models, and the results are shown in \ref{fig:robustnnes_NLU_erros}. All models achieved high performance with the dataset without errors. The RED, SEQ and TED achieved perfect performance. MD, MC, and PEDP also demonstrated high performance, although  slightly below than others. However, when increasing the errors to 10\%, we saw that all models experienced a decrease in their performance. In particular, the TED and RED models were the most affected, with a drop in performance to 80\%.

On the other hand, the SEQ model maintained the highest performance. As errors increased to 20\% and 40\%, the SEQ model showed the highest performance, closely followed by PEDP. RED, MD, MC, and TED continued to experience decreased performance, with TED being the most affected model. When errors reached 40\% and 60\%, the SEQ model showed notable robustness, maintaining its performance at 80\%. On the other hand, the performance of RED and TED fell significantly. Finally, even with very high error levels of 80\% and 90\%, the SEQ model showed remarkable robustness with a stable performance. In contrast, the other models experienced additional decreases in their performance, showing an almost linear trend.

In conclusion, our findings suggest that the TED, RED and SEQ models are notably robust when faced with datasets of varying complexity, maintaining high performance even on the most challenging datasets. The MD, MC, and PEDP models also demonstrated respectable performance, but they were more impacted by the increasing complexity of the datasets.  Importantly, these experiments also highlight that errors in datasets can significantly impact the performance of models, a factor that is often overlooked when comparing solutions. Our results show that the SEQ model proved to be the most resilient in the face of dataset errors, closely followed by PEDP. While all models experienced a performance drop with the introduction of errors, the SEQ model showed impressive robustness, maintaining consistent performance even at high error levels.
In contrast, the RED and TED models were significantly more impacted by the introduction of dataset errors. This study underscores the importance of considering dataset errors in model evaluation and comparison. Therefore, acknowledging the effects of errors in datasets is crucial for developing and deploying more reliable and efficient models.

\begin{table*}[!h]
\centering
\small
\begin{tabular}{@{}lcccccc@{}}
\toprule
\textbf{Models} & \multicolumn{3}{c}{\textbf{MultiWoz}} & \multicolumn{3}{c}{\textbf{SGD}} \\ 
\cmidrule(lr){2-4} \cmidrule(lr){5-7}
\textbf{} & \textbf{F1 \%} & \textbf{Precision \%} & \textbf{Recall \%} & \textbf{F1 \%} & \textbf{Precision \%} & \textbf{Recall \%} \\ 
\midrule
MC & 39.41 & 54.60 & 34.32 & 73.78 & 77.77 & 71.20 \\
MD & 35.92 & 51.93 & 30.10 & 78.37 & 90.33 & 72.32 \\
SEQ & 44.64 & 51.91 & 43.66 & \textbf{86.04} & 87.69 & \textbf{84.65} \\
RED & \textbf{69.52} & 65.27 & \textbf{69.52} & 74.44 & 74.27 & 77.61 \\
TED & 61.98 & 62.28 & 67.46 & 78.33 & 79.65 & 80.25 \\
PEDP & 66.95 & \textbf{78.11} & 65.02 & 84.74 & \textbf{92.07} & 81.30 \\
\bottomrule
\end{tabular}
\caption{Experimental results were obtained using all available datasets.}
\label{tab:all_datasets}
\end{table*}

\begin{table*}
\centering
\small
\begin{tabular}{lcccccccccc}
\toprule
\textbf{Models} & \multicolumn{3}{c}{\textbf{Simple}} & \multicolumn{3}{c}{\textbf{Medium}} & \multicolumn{3}{c}{\textbf{Hard}} \\ \cmidrule(lr){2-4} \cmidrule(lr){5-7} \cmidrule(lr){8-10}
& \textbf{F1\%} & \textbf{Precision\%} & \textbf{Recall\%} & \textbf{F1\%} & \textbf{Precision\%} & \textbf{Recall\%} & \textbf{F1\%} & \textbf{Precision\%} & \textbf{Recall\%} \\
\midrule
MC & 85,92 & 91,44 & 84,19 & 86,62 & 92,68 & 84,12 & 85,8 & 91,74 & 83,38 \\
MD & 81,91 & 89,72 & 80,19 & 80,25 & 90,31 & 77,66 & 80,45 & 90,36 & 77,87 \\
SEQ & \textbf{100} & \textbf{100} & \textbf{100} & \textbf{100} & \textbf{100} & \textbf{100} & \textbf{99,76} & \textbf{99,76} & \textbf{99,76} \\
RED & \textbf{100} & \textbf{100} & \textbf{100} & 98,9 & 98,99 & 98,95 & 90,11 & 94,97 & 89,55 \\
TED & 99,98 & 99,99 & 99,98 & 99,55 & 99,45 & 99,71 & 98,67 & 99,03 & 98,52 \\
PEDP & 84,85 & 91,27 & 81,5 & 83,08 & 95,95 & 76,57 & 87,55 & 97,45 & 81,56 \\
\bottomrule
\end{tabular}
\caption{Experimental results were obtained using the simple, medium, and hard synthetic datasets.}
\label{tab:combined}
\end{table*}

\section{Conclusion}

In this work, we have proved that dialogue management is a straightforward problem and that having a high-quality curated dataset is essential to evaluate the models correctly. Furthermore, we have proposed a taxonomy of the main errors in datasets, which need to be properly treated. In addition, we have designed a synthetic dataset generator that can become a helpful tool for anyone who wants to evaluate their dialog management models and test their dependence on errors, giving them a more detailed and deeper insight into the behaviour of their models.

\section*{Limitations}
Although our study yields findings and contributions to task-oriented dialogue systems, we must recognize certain limitations inherent in our approach.

First, although synthetic datasets are powerful for systematically controlling and examining errors, they cannot fully capture the complexity and variability of real-world data. Real dialogues are often imbued with subtleties, ambiguities, and nuances that can be difficult to replicate in a synthetic environment. Therefore, models that perform well on our synthetic datasets may not fully transfer their performance to real dialogues, limiting their practical applicability.

Second, it is essential to recognize that although error handling is a crucial factor in the performance of dialogue systems, it is not the only aspect that matters. The quality of natural language understanding and generation, the architecture of the model, and the choice of learning algorithm also play critical roles in system performance.

Finally, our proposal may face challenges in terms of scalability and complexity. As dialogue systems must handle an increasing diversity of tasks and domains, it may not be practical or feasible to generate synthetic datasets that cover all possible interactions.

Despite these limitations, our work marks a step forward in understanding the influence of data errors in task-oriented dialogue systems. Future work should explore these limitations and seek ways to overcome them further to improve the performance and robustness of these systems.

\section*{Ethics Statement}
Our approach to developing a synthetic dialogue generator raises particular ethical issues. Although the generator does not produce text and, therefore, cannot generate offensive or inappropriate content, the fact that it is used to generate data for training machine learning models raises other ethical considerations. For example, synthetic data may not reflect the diversity and complexity of fundamental human interactions in all their forms and contexts.

\section*{Acknowledgements}
ACIISI-Gobierno de Canarias and European FEDER Funds Grant EIS 2021 04 partially supported this research. 

\bibliography{emnlp2023}

\begin{thebibliography}{23}
\expandafter\ifx\csname natexlab\endcsname\relax\def\natexlab#1{#1}\fi

\bibitem[{Andreas et~al.(2020)Andreas, Bufe, Burkett, Chen, Clausman, Crawford,
  Crim, DeLoach, Dorner, Eisner, Fang, Guo, Hall, Hayes, Hill, Ho, Iwaszuk,
  Jha, Klein, Krishnamurthy, Lanman, Liang, Lin, Lintsbakh, McGovern,
  Nisnevich, Pauls, Petters, Read, Roth, Roy, Rusak, Short, Slomin, Snyder,
  Striplin, Su, Tellman, Thomson, Vorobev, Witoszko, Wolfe, Wray, Zhang, and
  Zotov}]{andreas-etal-2020-task}
Jacob Andreas, John Bufe, David Burkett, Charles Chen, Josh Clausman, Jean
  Crawford, Kate Crim, Jordan DeLoach, Leah Dorner, Jason Eisner, Hao Fang,
  Alan Guo, David Hall, Kristin Hayes, Kellie Hill, Diana Ho, Wendy Iwaszuk,
  Smriti Jha, Dan Klein, Jayant Krishnamurthy, Theo Lanman, Percy Liang,
  Christopher~H. Lin, Ilya Lintsbakh, Andy McGovern, Aleksandr Nisnevich, Adam
  Pauls, Dmitrij Petters, Brent Read, Dan Roth, Subhro Roy, Jesse Rusak, Beth
  Short, Div Slomin, Ben Snyder, Stephon Striplin, Yu~Su, Zachary Tellman, Sam
  Thomson, Andrei Vorobev, Izabela Witoszko, Jason Wolfe, Abby Wray, Yuchen
  Zhang, and Alexander Zotov. 2020.
\newblock \href {https://doi.org/10.1162/tacl_a_00333} {Task-oriented dialogue
  as dataflow synthesis}.
\newblock \emph{Transactions of the Association for Computational Linguistics},
  8:556--571.

\bibitem[{Bohus and Rudnicky(2009)}]{Bohus2009ravenClaw}
Dan Bohus and Alexander~I. Rudnicky. 2009.
\newblock The ravenclaw dialog management framework: Architecture and systems.
\newblock \emph{Comput. Speech Lang.}, 23(3):332–361.

\bibitem[{Brabra et~al.(2022)Brabra, Báez, Benatallah, Gaaloul, Bouguelia, and
  Zamanirad}]{surveyMarcos2022}
Hayet Brabra, Marcos Báez, Boualem Benatallah, Walid Gaaloul, Sara Bouguelia,
  and Shayan Zamanirad. 2022.
\newblock \href {https://doi.org/10.1109/TCDS.2021.3086565} {Dialogue
  management in conversational systems: A review of approaches, challenges, and
  opportunities}.
\newblock \emph{IEEE Transactions on Cognitive and Developmental Systems},
  14(3):783--798.

\bibitem[{Byrne et~al.(2019)Byrne, Krishnamoorthi, Sankar, Neelakantan,
  Goodrich, Duckworth, Yavuz, Dubey, Kim, and
  Cedilnik}]{byrne-etal-2019-taskmaster}
Bill Byrne, Karthik Krishnamoorthi, Chinnadhurai Sankar, Arvind Neelakantan,
  Ben Goodrich, Daniel Duckworth, Semih Yavuz, Amit Dubey, Kyu-Young Kim, and
  Andy Cedilnik. 2019.
\newblock \href {https://doi.org/10.18653/v1/D19-1459} {Taskmaster-1: Toward a
  realistic and diverse dialog dataset}.
\newblock In \emph{Proceedings of the 2019 Conference on Empirical Methods in
  Natural Language Processing and the 9th International Joint Conference on
  Natural Language Processing (EMNLP-IJCNLP)}, pages 4516--4525, Hong Kong,
  China. Association for Computational Linguistics.

\bibitem[{Eric et~al.(2020)Eric, Goel, Paul, Sethi, Agarwal, Gao, Kumar, Goyal,
  Ku, and Hakkani-Tur}]{eric-etal-2020-multiwoz}
Mihail Eric, Rahul Goel, Shachi Paul, Abhishek Sethi, Sanchit Agarwal, Shuyang
  Gao, Adarsh Kumar, Anuj Goyal, Peter Ku, and Dilek Hakkani-Tur. 2020.
\newblock \href {https://aclanthology.org/2020.lrec-1.53} {{M}ulti{WOZ} 2.1: A
  consolidated multi-domain dialogue dataset with state corrections and state
  tracking baselines}.
\newblock In \emph{Proceedings of the Twelfth Language Resources and Evaluation
  Conference}, pages 422--428, Marseille, France. European Language Resources
  Association.

\bibitem[{Gritta et~al.(2021)Gritta, Lampouras, and Iacobacci}]{GrittaCG2021}
Milan Gritta, Gerasimos Lampouras, and Ignacio Iacobacci. 2021.
\newblock \href {https://doi.org/10.1162/tacl_a_00352} {{Conversation Graph:
  Data Augmentation, Training, and Evaluation for Non-Deterministic Dialogue
  Management}}.
\newblock \emph{Transactions of the Association for Computational Linguistics},
  9:36--52.

\bibitem[{Grosz and Sidner(1986)}]{grosz-sidner-1986-attention}
Barbara~J. Grosz and Candace~L. Sidner. 1986.
\newblock \href {https://aclanthology.org/J86-3001} {Attention, intentions, and
  the structure of discourse}.
\newblock \emph{Computational Linguistics}, 12(3):175--204.

\bibitem[{Hochreiter and Schmidhuber(1997)}]{jurgen1997lstm}
Sepp Hochreiter and J\"{u}rgen Schmidhuber. 1997.
\newblock \href {https://doi.org/10.1162/neco.1997.9.8.1735} {Long short-term
  memory}.
\newblock \emph{Neural Comput.}, 9(8):1735–1780.

\bibitem[{Jang et~al.(2016)Jang, Gu, and Poole}]{jang2016gumbelsoftmax}
Eric Jang, Shixiang Gu, and Ben Poole. 2016.
\newblock \href {https://doi.org/10.48550/ARXIV.1611.01144} {Categorical
  reparameterization with gumbel-softmax}.

\bibitem[{Kim et~al.(2018)Kim, Song, Park, and Zunino}]{deleteNLU2018}
A-Yeong Kim, Hyun-Je Song, Seong-Bae Park, and Rodolfo Zunino. 2018.
\newblock \href {https://doi.org/10.1155/2018/5798684} {A two-step neural
  dialog state tracker for task-oriented dialog processing}.
\newblock \emph{Intell. Neuroscience}, 2018.

\bibitem[{Lee et~al.(2019)Lee, Zhu, Takanobu, Li, Zhang, Zhang, Li, Peng, Li,
  Huang, and Gao}]{convlad2019}
Sungjin Lee, Qi~Zhu, Ryuichi Takanobu, Xiang Li, Yaoqin Zhang, Zheng Zhang,
  Jinchao Li, Baolin Peng, Xiujun Li, Minlie Huang, and Jianfeng Gao. 2019.
\newblock \href {https://doi.org/10.48550/ARXIV.1904.08637} {Convlab:
  Multi-domain end-to-end dialog system platform}.

\bibitem[{Li et~al.(2020)Li, Kiseleva, and de~Rijke}]{li-etal-2020-rethinking}
Ziming Li, Julia Kiseleva, and Maarten de~Rijke. 2020.
\newblock \href {https://doi.org/10.18653/v1/2020.findings-emnlp.316}
  {Rethinking supervised learning and reinforcement learning in task-oriented
  dialogue systems}.
\newblock In \emph{Findings of the Association for Computational Linguistics:
  EMNLP 2020}, pages 3537--3546, Online. Association for Computational
  Linguistics.

\bibitem[{Mosig et~al.(2020)Mosig, Vlasov, and Nichol}]{Mosing2020context}
Johannes E.~M. Mosig, Vladimir Vlasov, and Alan Nichol. 2020.
\newblock \href {https://doi.org/10.48550/ARXIV.2004.10473} {Where is the
  context? -- a critique of recent dialogue datasets}.

\bibitem[{Ni et~al.(2022)Ni, Young, Pandelea, Xue, and Cambria}]{surveyNi2022}
Jinjie Ni, Tom Young, Vlad Pandelea, Fuzhao Xue, and Erik Cambria. 2022.
\newblock \href {https://doi.org/10.1007/S10462-022-10248-8} {Recent advances
  in deep learning based dialogue systems: a systematic survey}.
\newblock \emph{Artificial Intelligence Review 2022}, pages 1--101.

\bibitem[{Rastogi et~al.(2020)Rastogi, Zang, Sunkara, Gupta, and
  Khaitan}]{rastogi2020sgd}
Abhinav Rastogi, Xiaoxue Zang, Srinivas Sunkara, Raghav Gupta, and Pranav
  Khaitan. 2020.
\newblock Towards scalable multi-domain conversational agents: The
  schema-guided dialogue dataset.
\newblock In \emph{Proceedings of the AAAI Conference on Artificial
  Intelligence}, volume~34, pages 8689--8696.

\bibitem[{Vlasov et~al.(2018)Vlasov, Drissner-Schmid, and
  Nichol}]{Vlasov2018REPD}
Vladimir Vlasov, Akela Drissner-Schmid, and Alan Nichol. 2018.
\newblock \href {https://doi.org/10.48550/ARXIV.1811.11707} {Few-shot
  generalization across dialogue tasks}.

\bibitem[{Vlasov et~al.(2019)Vlasov, Mosig, and Nichol}]{Vlasov2019}
Vladimir Vlasov, Johannes E.~M. Mosig, and Alan Nichol. 2019.
\newblock \href {https://doi.org/10.48550/ARXIV.1910.00486} {Dialogue
  transformers}.

\bibitem[{Weizenbaum(1966)}]{Weizenbaum1966}
Joseph Weizenbaum. 1966.
\newblock \href {https://doi.org/10.1145/365153.365168} {Eliza—a computer
  program for the study of natural language communication between man and
  machine}.
\newblock \emph{Commun. ACM}, 9(1):36–45.

\bibitem[{Wu et~al.(2018)Wu, Fisch, Chopra, Adams, Bordes, and
  Weston}]{Wu_Fisch_Chopra_Adams_Bordes_Weston_2018}
Ledell Wu, Adam Fisch, Sumit Chopra, Keith Adams, Antoine Bordes, and Jason
  Weston. 2018.
\newblock \href {https://doi.org/10.1609/aaai.v32i1.11996} {Starspace: Embed
  all the things!}
\newblock \emph{Proceedings of the AAAI Conference on Artificial Intelligence},
  32(1).

\bibitem[{Ye et~al.(2022)Ye, Manotumruksa, and
  Yilmaz}]{ye-etal-2022-multiwoz2.4}
Fanghua Ye, Jarana Manotumruksa, and Emine Yilmaz. 2022.
\newblock \href {https://aclanthology.org/2022.sigdial-1.34} {{M}ulti{WOZ} 2.4:
  A multi-domain task-oriented dialogue dataset with essential annotation
  corrections to improve state tracking evaluation}.
\newblock In \emph{Proceedings of the 23rd Annual Meeting of the Special
  Interest Group on Discourse and Dialogue}, pages 351--360, Edinburgh, UK.
  Association for Computational Linguistics.

\bibitem[{Zhang et~al.(2022)Zhang, Zhao, Wang, Li, Huang, and Feng}]{pepd2022}
Shuo Zhang, Junzhou Zhao, Pinghui Wang, Yu~Li, Yi~Huang, and Junlan Feng. 2022.
\newblock \href {https://doi.org/10.48550/ARXIV.2204.11481} {"think before you
  speak": Improving multi-action dialog policy by planning single-action
  dialogs}.

\bibitem[{Zhang et~al.(2020)Zhang, Takanobu, Zhu, Huang, and
  Zhu}]{surveyZhang2020}
Zheng Zhang, Ryuichi Takanobu, Qi~Zhu, MinLie Huang, and XiaoYan Zhu. 2020.
\newblock \href {https://doi.org/10.1007/s11431-020-1692-3} {Recent advances
  and challenges in task-oriented dialog systems}.
\newblock \emph{Science China Technological Sciences}, 63(10):2011--2027.

\bibitem[{Zhang et~al.(2019)Zhang, Li, Gao, and
  Chen}]{zhang-etal-2019-budgeted}
Zhirui Zhang, Xiujun Li, Jianfeng Gao, and Enhong Chen. 2019.
\newblock \href {https://doi.org/10.18653/v1/P19-1364} {Budgeted policy
  learning for task-oriented dialogue systems}.
\newblock In \emph{Proceedings of the 57th Annual Meeting of the Association
  for Computational Linguistics}, pages 3742--3751, Florence, Italy.
  Association for Computational Linguistics.

\end{thebibliography}
\bibliographystyle{acl_natbib}

\end{document}